\documentclass{article}
\usepackage{spconf,amsmath,graphicx,amsfonts}

\title{FACIAL MICRO-EXPRESSION SPOTTING AND RECOGNITION USING TIME
CONTRASTED FEATURE WITH VISUAL MEMORY}
\name{Sauradip Nag \textsuperscript{1}, Ayan Kumar Bhunia \textsuperscript{2}\sthanks{Corresponding Author}, Aishik Konwer \textsuperscript{3}, Partha Pratim Roy \textsuperscript{4}}
\address{\textsuperscript{1}Kalyani Government Engineering College, India \hspace{0.1cm} \textsuperscript{2}Nanyang Technological University, Singapore\\ \textsuperscript{3} Cognizant Technology Solutions, India \hspace{0.1cm} \hspace{0.1cm} \textsuperscript{4}Indian Institute of Technology Roorkee, India \\
{\tt\small \textsuperscript{2}ayanbhunia@ntu.edu.sg }
}
\begin{document}
%
\maketitle
\begin{abstract}
Facial micro-expressions are sudden involuntary  minute muscle movements which reveal true emotions that people try to conceal. Spotting a micro-expression and recognizing it is a major challenge owing to its short duration and intensity. Many works pursued traditional and deep learning based approaches to solve this issue but compromised on learning low level features and higher accuracy due to unavailability of datasets. This motivated us to propose a novel joint architecture of spatial and temporal network which extracts time-contrasted features from the feature maps to contrast out micro-expression from rapid muscle movements. The usage of time contrasted features greatly improved the spotting of micro-expression from inconspicuous facial movements. Also, we include a memory module to predict the class and intensity of the micro-expression across the temporal frames of the micro-expression clip. Our method achieves superior performance in comparison to other conventional approaches on CASMEII  dataset.  
\end{abstract}
\section{Introduction}
\label{sec:intro}

Expression Recognition is a widely studied topic in the domain of Pattern Recognition and Biometrics in the past few decades. In addition to verbal communication there exists some nonverbal communication such as facial expression which reveals true emotional state of a human being. Facial expression is a result of muscle contractions of the face which helps in representing wide range of emotions such as disgust, sadness, happiness, fear, anger and surprise. Facial expressions can be broadly classified into three categories based on duration, motion range and action areas namely (a) Macro Expressions (b) Micro Expressions and (c) Subtle Expressions.

Micro-expressions are hidden emotions and occur over small region of the face which is a major hurdle. Motivated by this challenge, recent works have been focused mainly on the hurdles of spotting micro-expressions like hand crafted feature extraction, dataset creation and micro-expression classification.

Based on feature extraction, recent works on Micro-expressions can be broadly classified into two main categories; namely (1) Handcrafted Feature Based Methods (2) Deep Learning Based Approaches. Handcrafted based approaches include feature extraction techniques such as 3D Histograms of Oriented Gradients (3DHOG), Local Binary Pattern - Three Orthogonal Plane (LBP-TOP) and Histogram of Optical Flow (HOOF). In the works of micro-expression recognition using 3DHOG. Chen et al. \cite{chen2016emotion} proposed 3DHOG features with weighted method and used fuzzy classification for recognition of micro-expression. Pfister et al. \cite{pfister2011recognising} used LBP-TOP as a spatio-temporal local texture descriptor to extract dynamic features and used the combination of Support Vector Machine (SVM), Multi Kernel Learning (MKL), Random Forest (RF) as classifiers for the classification of spontaneous facial micro-expressions. In another work, Pfister et al. \cite{pfister2011differentiating} extended Complete Local Binary Pattern (CLBP) to work with dynamic texture descriptor and called it CLBP-TOP and used the mix of SVM, linear classifier, and Random Forest as classifiers for recognition.
Liu et al. \cite{liu2016main} proposed Main Directional Mean Optical Flow (MDMO) as a feature for recognition of micro-expression. They used RF as classifier which achieved better accuracy than previous benchmarks. 
Happy and RoutRay \cite{happy2017fuzzy} proposed temporal based Fuzzy Histogram of Optical Flow Orientation (FHOFO) as feature descriptor to focus on the facial changes during micro-expression.
In the last few years, Deep Learning based approaches has rapidly grown and made its way into a major number of computer vision and biometric problems which replaced the age-old traditional algorithms and hand-crafted feature extraction methods. With the availability of large sized datasets like Sports 1M Dataset \cite{karpathy2014large}, training these Neural Networks are quite easy. However, only few of the works on micro-expression consider Deep Learning pipeline due to the fact that the micro-expression databases like (i) CASME \cite{yan2013casme} (ii) CASME2 \cite{yan2014casme} and CAS(ME)\textsuperscript{2} \cite{qu2017cas} contain a collection of only 495 Micro-Expression Clips which is insufficient to train a CNN.  Kim et al. \cite{kim2016micro} was the first to integrate CNNs into micro-expression where they encoded a feature representation of temporal states like onset, apex and offset using a CNN. Then these encoded features were passed into Long Short Term Memory (LSTM), where temporal characteristics were analyzed. In 2017, Peng et al. \cite{peng2017dual} proposed a dual stream based CNN Network called Dual Temporal Scale CNN (DTSCNN) combined with optical flow.
Deep Learning based approaches improved the recognition accuracy but it suffered from learning low level features due to unavailability of training data. But these are very essential in the context of spotting micro-expression. In this paper, we proposed a novel approach that aims to address these drawbacks. To the best of our knowledge, this is the first work to deal with the micro-expression spotting problem by employing context contrasted feature and adding memory element to recognize intensity and class.
 
The contributions of this work are as follows: firstly, we extracted both Spatial and Temporal Information using a joint architecture for building a strong representation of micro-expression features. Secondly, we introduced context contrasted features for high level feature modeling to contrast micro-expressions from the relevant muscle movements. Finally, Memory Element in the form of GRU is used to encode the spatio-temporal evolution of micro-expression and to improve the expression class separability of the learned micro-expression features.
The rest of the paper is organized as follows: in Section 2, we discuss the proposed approach in details. In section 3, we describe the results and experiment results. Finally, the conclusion is provided. 
\vspace{-0.3cm}

\section{Proposed Method}
\label{sec:format}
Our framework for spotting and recognition of facial micro-expression takes both low and high level features into consideration. Also, efficacy is not compromised despite having less available data. The proposed model consists of 4 main components namely: Spatial Network, Temporal Network, Context-Contrast Module and Visual Memory Module. The detailed architecture is represented in Fig 1. 

%

\begin{figure}[h]
\label{pic1}
\includegraphics[width=8.9cm , height = 7.1cm , ]{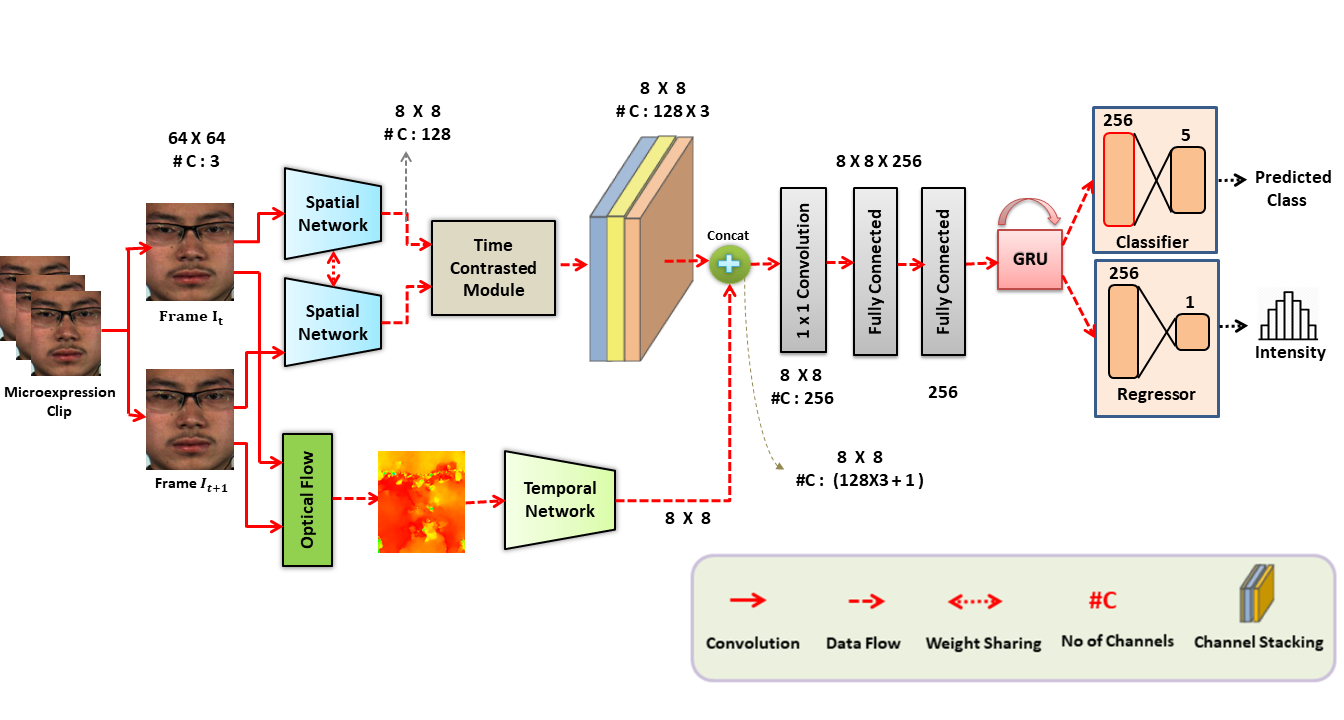}
\caption{Illustration of the proposed framework: It consists of two networks Spatial Network and Temporal Network. Spatial and temporal network takes a temporal frame $I_{t}$ and subsequent frame $I_{t+1}$ and tries to generate spatial feature maps for the frames from which time constrasted module contrasts out the feature relevant to micro-expression. The memory module takes this contrasted feature to predict the class and intensity of the micro-expression
 }
  \label{figure:1}
\end{figure}
\vspace{-0.2cm}
\subsection{ Micro-Expression Spotting }
\label{ssec:subhead}
    
Micro-expression is a rapid muscle movement within a fraction of second. Spotting them in temporal videos in spite of variable lighting conditions and head movements is quite a difficult task. To address this problem, we propose a joint architecture considering temporal and spatial information. With the help of this joint architecture it is also possible to identify the start and end frame of the micro-expression. We also introduced Context-Contrast Module which enhances or contrasts away the micro-expression from the visible background/foreground objects which is discussed in the following subsections.

\textit{Spatial Network: }The primary aim of Spatial Network is to learn the high level spatial features which are extracted in the form of edges. This will help to produce a higher-level encoding of micro-expression image for assisting GRU in the formation of representation of minute muscle movements. In every iteration,  the network takes 2 subsequent temporal frames I\textsubscript{t} and I\textsubscript{t+1} of a video having dimensions  $ 64 $ $ \times{64} $ $ \times{3} $  as input as shown in fig 1. Each of the temporal frame produces a $ 128 $ $ \times{\frac{w}{8}} $ $ \times{\frac{h}{8}} $ feature representation. As a result, the semantic content of the micro-expression is encoded. The architecture used in this module is the state-of-the-art CNN for this channel, namely the  DeepLab largeFOV \cite{chen2014semantic}. This network preserves a relatively high spatial resolution of features and also incorporates context information in each pixel’s representation. The network has been trained on CASMEII dataset, hence it can segment out muscular features related to micro-expression with ease. The features extracted from the fc6 layer of the network, have dimension of 1024 for each pixel. This feature map is further passed through two 1 x 1 convolutional layers, interleaved with tanh nonlinearities, to reduce the dimension to 128. These layers are trained together with GRU.

\textit{Temporal Network: }For the extraction of High level temporal information per pixel, we employed a temporal Network. The underlying architecture of this network is MPNet \cite{tokmakov2017learning} which is a CNN pretrained model for the motion segmentation task. MPNet or Motion Pattern Network is a encoder-decoder style architecture which aims to segment the motion pattern in flow fields. MPNet has been trained to estimate independently moving micro-expression (i.e. irrespective of camera motion) based on optical flow computed from a pair of frames as input. A $ 64 $ $ \times{64} $ matrix having optical flow information between two frames, is passed to the network as input. The output has $ \frac{w}{4} $ $ \times{\frac{h}{4}} $ dimension, where each value represents the likelihood of the corresponding pixel being in motion. This output is further downsampled by a factor 2 (in w and h) to match the dimensions of the spatial network output since our aim is to incorporate both temporal and spatial features. The benefit of using two networks is that both spatial and temporal networks are equally effective when a facial muscle is moving in the micro-expression clip. But as soon as it becomes stationary, the temporal network can not estimate the object, unlike the spatial network. 

\textit{Context Constrast Module: }
When human beings look at one object, they tend to concentrate on that object while they are aware of the surroundings. Similar concept has been formulated to create a context constrast module as described in \cite{ding2018context}. The feature of this module is that it enhances the micro-expression and in spite of visible noise, contrasts the micro-expression from the foreground. It is very hard to collect appropriate and discriminating high level features for a particular pixel or patch. Hence it can be solved by generating local information and context separately, and then fusing them after making contrast between these two informations. For spotting the micro-expression, we perform intra-map difference between the feature maps coming out of Spatial Network for frames I\textsubscript{t} and I\textsubscript{t+1}. The contrasting features are taken based on local and global context of both input temporal frames using the following equations : 

\vspace{-0.42cm}
\begin{equation}
F\textsubscript{L1-C2} = M\textsubscript{l}(FM\textsubscript{t} )- M\textsubscript{c}(FM\textsubscript{t+1} )
\end{equation}
\begin{equation}
F\textsubscript{L1-L2} = M\textsubscript{l}(FM\textsubscript{t} )- M\textsubscript{l}(FM\textsubscript{t+1}  )
\end{equation}
\begin{equation}
F\textsubscript{L2-C1} = M\textsubscript{l}(FM\textsubscript{t+1} )- M\textsubscript{c}(FM\textsubscript{t})
\end{equation}
where FM\textsubscript{t} is the feature map of frame t among the temporal frame,  M\textsubscript{l} is the function computing local feature of frame,  M\textsubscript{c} is the function computing context feature of frame and output F is the frame difference of the local and global features. Next, the frame-difference feature maps are stacked together and fused to the existing feature maps of the spatial network. Finally the context-contrasted spatial features and temporal features are concatenated for micro expression recognition step.

%

\subsection{ Micro-Expression Recognition }
\label{ssec:subhead}

For micro-expression recognition, feature extraction is a critical issue as discussed earlier. In recent works, analysis of spontaneous facial micro-expression  has been receiving attention from numerous researchers since involuntary micro-expressions can reveal genuine emotions which people try to conceal. Hence for micro-expression recognition, the memory module plays an important role since it processes sequential data and also because ordering of frame is a vital part of recognition and spotting.

\textit{Memory Module: }
The reason behind choosing GRU as the memory module in this framework is because it is a special form of lightweight RNN having memory cell which can remember long-term information. We have used GRU to decode the micro-expression from temporal frames to predict expression-class and expression-intensity by taking past time-step information into consideration. The depth of the concatenated feature maps of both spatial and temporal network is first reduced using a $ 1 $ $  \times{1} $ convolution layer to a dimension of $ 256 $. The output of this $ 1 \times 1 $ convolution operation is then flattened using 2 Fully connected Layer to obtain a dimension of $ 256 \times 1$. This vector is given as input to the GRU. 
The output of the GRU has two heads: 1) It provides the likelihood of the expression class which is then passed through a softmax function to predict the class, 2) The expression-state intensity of the micro-expression to guess the start and end point of the micro-expression. However, note that, uni-directional GRU is chosen in our work because the order of the features for recognition is very important. 
The classification head and Regression heads are fully connected layers of dimension  $\mathbb{R}^{256 \times 1}$. The fully connected layer of classification head is further passed into another fully connected layer of dimension  $\mathbb{R}^{5 \times 1}$, since CASME2 dataset contains 5 classes of micro-expression. Thus, we obtain a 5-class vectorized output. It is then fed to the softmax regression to predict the correct class. The loss function used for classification head is Cross Entropy Loss
\vspace{-0.3cm}
\begin{equation}
L_{class} = -\dfrac{1}{N}\sum_{i=1}^{n}\langle y_{true}(x_{i} , y_{i} )\log( y_{pred}(x_{i},y_{i}) ) \rangle
\end{equation}
where N is the number of observations $ y_{true}(x_{i} , y_{i}) $ is the distribution of training data  and $ y_{pred}(x_{i} , y_{i}) $ is the distribution which comes out of the softmax layer. Similarly, the fully connected layer of regression head is mapped to 1 neuron which is passed through a sigmoid activation. The loss function used for regression is L1 loss. We obtain the loss by computing the L1 distance between the true intensity $ {Y_{int}} $ and model intensity $ \widehat{y_{int}} $ 
\vspace{-0.2cm}
\begin{equation}
L_{reg} = \dfrac{1}{N}\sum_{i=1}^{n}\Vert \widehat{y_{int}(i)} - {Y_{int}(i)} \Vert_{1}
\end{equation}
The output of classifier are micro-expression classed and that of regression is state of expression. The state is further classified into 6 groups depending on the value of state namely onset, onset-apex, apex, apex-offset, offset.

\vspace{-0.2cm}
\section{Experiments and Results}
\label{sec:pagestyle}

In this section we present result and analysis of the proposed method.

\textbf{Database: } For evaluating the efficiency of the proposed approach, CASMEII \cite{yan2014casme} dataset has been used to train all the baseline architectures. The temporal and spatial networks are trained separately. After that, the trained model is used to train the intermediate Convolution Layers and GRU Module. This dataset is chosen because it contains samples which are spontaneous, and dynamic micro-expressions with high frame rate (200 fps). The dataset contains 247 micro-expressions labeled with action units (AUs). But the number of samples are insufficient for training deep learning models. Hence, we need to augment dataset to increase the number of samples and also prevent overfitting of the model. Each training sequence was rotated in the range of [ $-10^{\circ}$ , $ 10^{\circ} $ ], scaled in the range of [ 0.9 , 1.1 ] and translated in the range of [ -2 to +2 ] along both the axes. Hence, 150 augmented sequence of each training sequence is generated, which substantially increased the current dataset size and made it suitable for training.

\textbf{Training: } The temporal and spatial networks used in the architecture have been trained separately using CASMEII Dataset following the parameters as mentioned in \cite{chen2014semantic}, \cite{tokmakov2017learning}. After that, the trained weights are used to train the memory module by minimizing 2 Loss-heads at each timestep of GRU. These are binary cross-entropy loss for Classification task and L1 Loss for Regression task. We used $ 70 \% - 30 \% $ train-test split and leave-one-out cross validation. Weight update is done using back-propagation through time and RMSProp optimizer with a learning rate of $ 10^{-4}$ and the weight decay is set to 0.005. Initialization of all the convolutional layers, except for those inside the GRU, is done with the standard Xavier method. The batch size is set to  14 where a sequence of 14 frames are chosen after randomly selecting a video. The entire GRU module uses $ 7 \times 7 $ convolutions in all the operations. The weights of two $1 \times 1$ convolutional (for dimensionality reduction) layers in the spatial network, two fully connected layers and one $ 1 \times 1$ Convolution (for dimensionality reduction) preceeding the memory module, are learned jointly with the memory module. Experiments are conducted on a server with 12 GB memory and single Nvidia Tesla K80 GPU. The model implementation is done using Tensorflow library.

\textbf{Results:} We compare our proposed method with some of the existing deep learning approaches, traditional approaches and baseline methods performed on CASMEII Dataset for both Micro-expression spotting and recognition. The metrics used for micro-expression spotting are evaluated using receiver operating characteristic (ROC) curves. The area under ROC, known as AUC, is used as an evaluation metric. The proposed method is compared with traditional LBP and HOOF feature extraction techniques which are performed on CASMEII Dataset as shown in table \ref{tab:table-1}. 
\begin{table}[h]
\centering
\begin{tabular}{|c|c|}
\hline
\textbf{Techniques} & \textbf{AUC}      \\ \hline
LBP                 & 92.98\%           \\ \hline
HOOF                & 64.99 \%          \\ \hline
\textbf{Proposed}       & \textbf{94.24 \%} \\ \hline
\end{tabular}

\caption{\label{tab:table-1}Quantitative Results of  micro-expression spotting on CASMEII Dataset }
\end{table}
From the results in Table \ref{tab:table-1} it is evident that deep learning based feature extraction outperforms traditional handcrafted feature extraction. 

\begin{table}[h]
\centering
\begin{tabular}{|c|c|}
\hline
\textbf{Methods}  & \textbf{Accuracy} \\ \hline
Baseline \cite{yan2014casme}         & 63.41 \%          \\ \hline
Kim et al. \cite{kim2016micro}         & 60.98 \%          \\ \hline
Peng et al. \cite{peng2017dual}        & 66.67 \%          \\ \hline
Adaptive MM \cite{wang2017effective}      & 75.30 \%          \\ \hline
Park et al. \cite{park2015subtle}        & 69.63 \%          \\ \hline
\textbf{Proposed} & \textbf{81.46 \%} \\ \hline

\end{tabular}
\caption{\label{tab:table-2}Quantitative Results of  micro-expression recognition on CASMEII Dataset }
\end{table} 
From the values of Table \ref{tab:table-2}, it is evident that our approach achieved superior performance compared to most of the traditional approaches and baseline methods due to the inclusion of time contrasted feature within the feature maps. However, the accuracy can be improved even further with the improvement of dataset and increasing the size of actual data instead of synthetic data generation.
The qualitative results are shown as in Fig 2 where the frame histogram can be seen as 
  
\begin{figure}[htb]

\begin{minipage}[b]{1.0\linewidth}
\centerline{\includegraphics[height= 3.4cm , width=\linewidth]{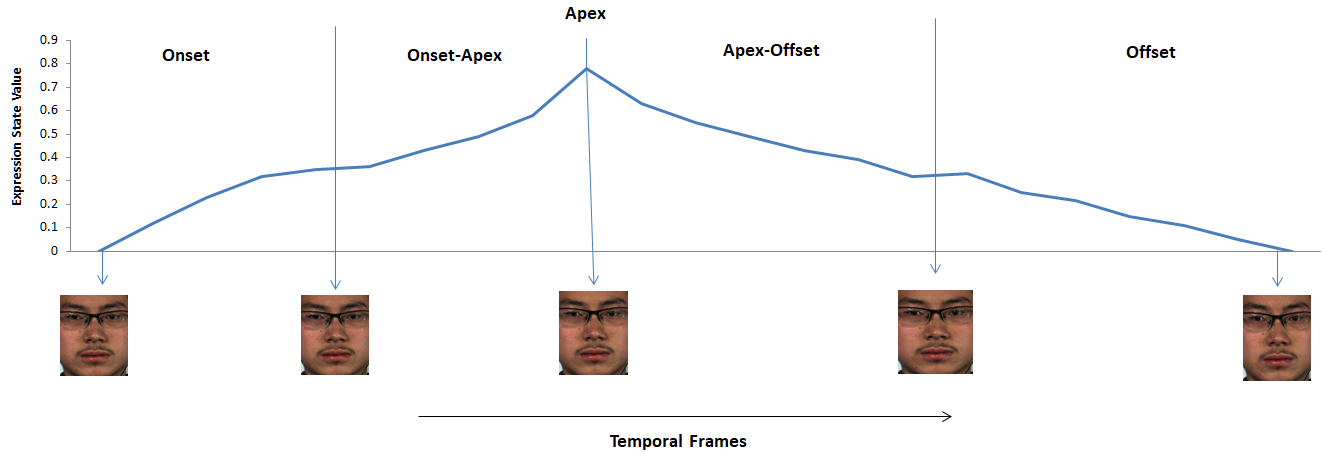}}
  \vspace{-0.1cm}
\caption{Qualitative results of our proposed method: the intensity of the micro-expression predicted for each temporal frame is grouped into different expression-states(onset,onset-apex,apex,apex-offset,offset)    }\medskip
\end{minipage}
\end{figure}
\vspace{-0.8cm}
\section{Conclusion}
In this paper we addressed the micro-expression spotting and recognition problem by the inclusion of a time contrasted feature and a memory module in our novel approach. The time contrasted feature contrasts the micro-expression context from the higher level spatial feature maps given by Spatial Network. This amplifies the representation of micro-expressions from other muscle movements. The GRU module added a memory element within the architecture that remembers and predicts the flow of micro-expression intensity across the temporal frames of the micro-expression clip. The experimental results, carried out on CASMEII dataset, suggests that the proposed method outperforms other existing approaches and baseline methods in terms of recognition rate.

\vfill\pagebreak

%

\bibliographystyle{IEEEbib}
\bibliography{refs}

\end{document}